\documentclass{article}
\usepackage{ijcai26}
\usepackage{xcolor}
\usepackage{times}
\usepackage{soul}
\usepackage{url}
\usepackage[hidelinks]{hyperref}
\usepackage[utf8]{inputenc}
\usepackage[small]{caption}
\usepackage{graphicx}
\usepackage{amsmath}
\usepackage{amsthm}
\usepackage{booktabs}
\usepackage{algorithm}
\usepackage{algorithmic}
\usepackage[switch]{lineno}
\usepackage{booktabs}
\usepackage[table]{xcolor}
\usepackage{pifont}
\usepackage{tcolorbox}
\usepackage{pifont}
\newcommand{\cmark}{\ding{51}} 
\newcommand{\xmark}{\ding{55}} 
\usepackage{multirow}
\usepackage{arydshln}

\title{Property Enhanced Instruction Tuning for Multi-Task Molecule Generation with Large Language Models}

\author{
    Xuan Lin$^1$\and
    Long Chen$^1$\and
    Yile Wang$^2$\thanks{Corresponding author.}\and
    Yangyang Chen$^3$\And
    Xiangxiang Zeng$^4$\\
    \affiliations
    $^1 $School of Computer Science, Xiangtan University\\
    $^2 $College of Computer Science and Software Engineering, Shenzhen University\\
    $^3 $ Department of Computer Science, University of Tsukuba\\
    $^4 $ College of Computer Science and Electronic Engineering, Hunan University\\
    \emails
    wangyile@szu.edu.cn
}

\begin{document}

\maketitle

\begin{abstract}
    Large language models (LLMs) are widely applied in various natural language processing tasks such as question answering and machine translation. However, due to the lack of labeled data and the difficulty of manual annotation for biochemical properties, the performance for molecule generation tasks is still limited, especially for tasks involving multi-properties constraints. In this work, we present a two-step framework PEIT (\textbf{P}roperty \textbf{E}nhanced \textbf{I}nstruction \textbf{T}uning) to improve LLMs for molecular-related tasks. In the first step, we use textual descriptions, SMILES, and biochemical properties as multimodal inputs to pre-train a model called PEIT-GEN, by aligning multi-modal representations to synthesize instruction data. In the second step, we fine-tune existing open-source LLMs with the synthesized data, the resulting PEIT-LLM can handle molecule captioning, text-based molecule generation, molecular property prediction, and our newly proposed multi-constraint molecule generation tasks. Experimental results show that our pre-trained PEIT-GEN outperforms MolT5, BioT5, MolCA and Text+Chem-T5 in molecule captioning, demonstrating modalities align well between textual descriptions, structures, and biochemical properties. Furthermore, PEIT-LLM shows promising improvements in multi-task molecule generation, demonstrating the effectiveness of the PEIT framework for molecular tasks. The code and appendix are available at \url{https://github.com/chenlong164/PEIT}.
\end{abstract}


\begin{figure}[t!]
	\centering
	\includegraphics[scale=1]{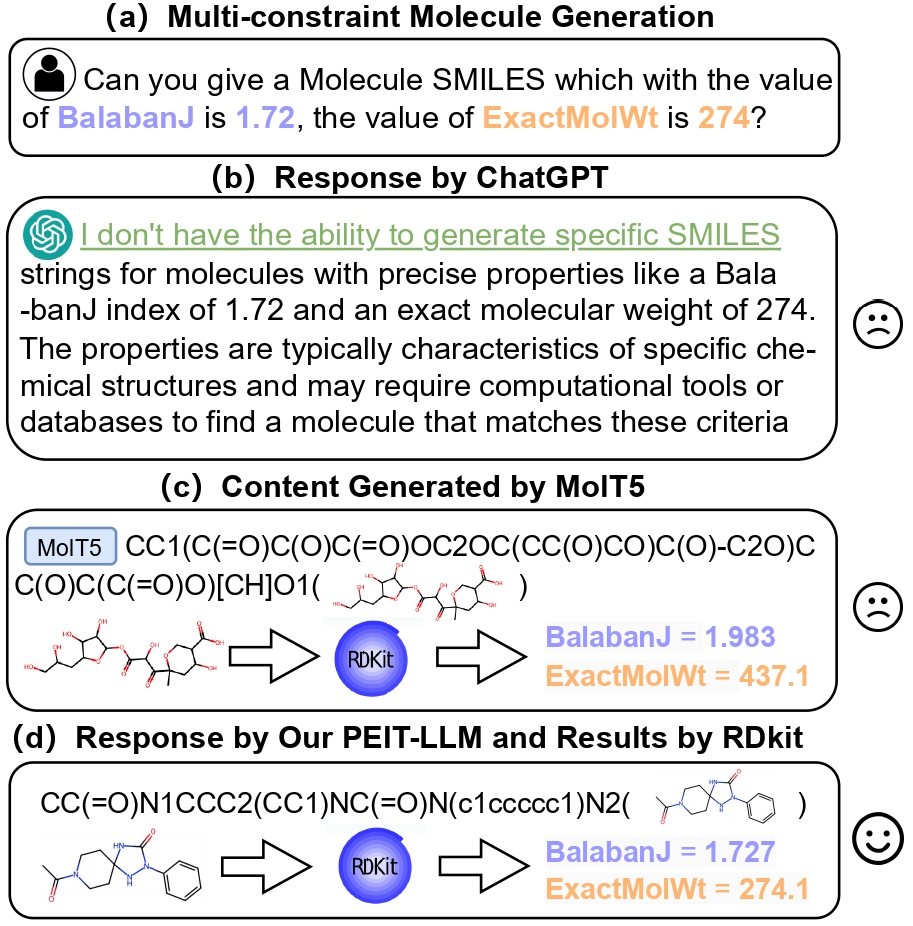}
	\caption{(a) An example of our proposed multi-constraint molecule generation task. (b) The response by ChatGPT. (c) The result generated by MolT5. (d) The response generated by the LLaMA3.1 model after applying our proposed property-enhanced instruction tuning, with the results validated by RDKit.}
	\label{figure:intro}
\end{figure}

\section{Introduction}

Large language models (LLMs) such as GPT-4~\cite{achiam2023gpt}, Gemini~\cite{comanici2025gemini} and DeepSeek~\cite{guo2025deepseek} have revolutionized the landscape of artificial intelligence and natural language processing, allowing machines to understand and generate human language with remarkable fluency and coherence. Based on encoded world knowledge~\cite{petroni2019language} and powerful instruct-following~\cite{zhang2023instruction} capabilities of LLMs, recent work has successfully used LLM for molecular-related tasks, achieving promising results~\cite{cao2025instructmol}. 

Despite their success, as illustrated in Figure~\ref{figure:intro} (a) and (b), LLMs still struggle with generating molecules under strict property constraints. Even specialized molecular translation models, such as MolT5~\cite{edwards-etal-2022-translation}, fail in these tasks (see Figure~\ref{figure:intro} (c)). While these models effectively capture relationships between molecular text and structure, they lack sufficient understanding of molecular properties, limiting their ability to incorporate property constraints in prompts. This shortcoming restricts their practical utility in applications like drug discovery~\cite{zhavoronkov2018artificial}.

The challenges in addressing such tasks mainly lie in three aspects: (1) Existing studies have revealed the limitations of LLMs in understanding molecular representations~\cite{grisoni2023chemical}, making it difficult to handle tasks requiring precise molecular property comprehension; (2) While there are some known SMILES-property pairing data, it often remains limited to predicting a single property and lacks datasets that cover a wide range of properties ~\cite{wu2018moleculenet}. Moreover, most of these datasets do not include precisely described textual data, making it challenging to identify accurate tri-modal data pairs; (3) To our knowledge, there are no suitable datasets or evaluation methods exist for multi-constraint molecule generation using LLMs, which challenges the standardization and assessment of such tasks~\cite{elton2019deep}.


\begin{figure*}[t!]
    \centering
    \includegraphics[width=0.95\linewidth]{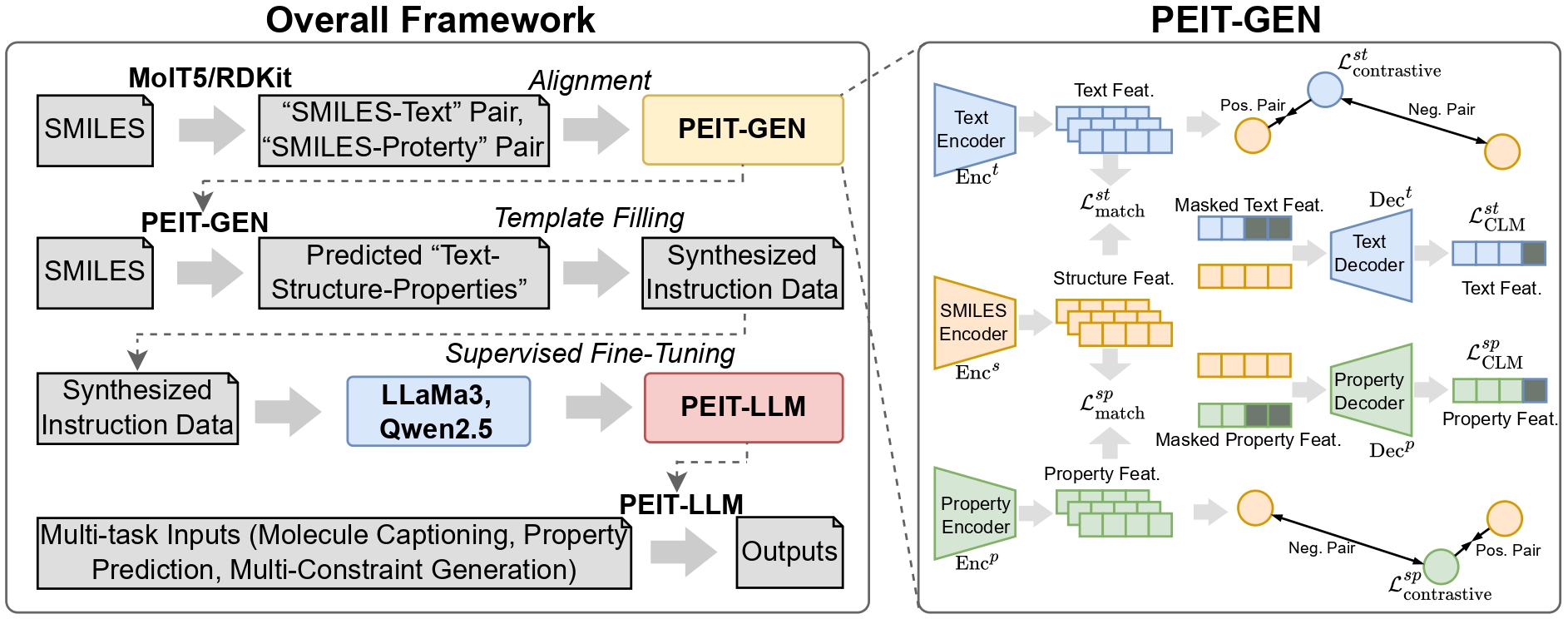}
    \caption{Left: Overall PEIT framework. We first pre-train the PEIT-GEN and construct instruction data via template filling. Then we fine-tune the open-source LLMs through instruction tuning, the resulting PEIT-LLM is used for multi-task molecule generation. Right: The process of PEIT-GEN pre-training, see details in Section~\ref{sec:peit-gen}.}
    \label{figure:framework}
\end{figure*}
To address these challenges, we propose a framework called PEIT (\textbf{P}roperty \textbf{E}nhanced \textbf{I}nstruction \textbf{T}uning) to generate multi-modal molecular instruction datasets in bulk, aiming to enhance capabilities of LLMs in multi-task molecule generation. \textcolor{black}{Using the} PEIT framework, our pre-trained model can handle general tasks (e.g., molecule captioning~\cite{edwards-etal-2022-translation}) and property-related tasks such as property prediction~\cite{chang2024bidirectional}. This makes it suitable for constructing data to evaluate multi-constraint molecule generation capabilities and for serving as instruction tuning data to improve existing open-source LLMs.

The overall structure of the proposed PEIT framework is shown in the left of Figure~\ref{figure:framework}. Specifically, it consists of two components: (1) We pre-train a model called PEIT-GEN through multi-modal representation alignment, which integrates text-based (molecular descriptions), structure-based (SMILES), and property-based (property-value pairs) information to generate diverse unstructured text, sequence, and property data; (2) By using the synthesized instruction data, we fine-tune open-source LLMs and develop PEIT-LLM, which can be applied to various molecule generation tasks mentioned above, including our \textcolor{black}{proposed} multi-constraint molecule generation.

Experimental results demonstrate that our pre-trained PEIT-GEN achieves competitive or better results in molecule captioning tasks, comparing to a variety of biomolecular models including MolT5~\cite{edwards-etal-2022-translation}, BioT5~\cite{pei2023biot5}, GIT-Mol~\cite{liu2024git}, MolXPT~\cite{liu2023molxpt}, MolCA~\cite{liu2023molca}, and Text+Chem-T5~\cite{christofidellis2023unifying}. Additionally, PEIT-LLM based on LLaMa3.1-8B~\cite{dubey2024llama} exhibits superior performance compared to specialized \textcolor{black}{models} Mol-Instructions~\cite{fang2023mol} and general-purpose LLMs including LLaMa3~\cite{dubey2024llama} and Qwen2.5~\cite{Qwen2TechnicalReport} in molecular property prediction and our newly proposed multi-constraint molecule generation tasks. 

\section{Related Work}

\textbf{Molecule Generation.} Molecule generation tasks fall into two categories: (1) text-based molecule generation that uses textual descriptions to generate molecules that match the given description~\cite{liu2023molxpt,liu2024git}. MolT5~\cite{edwards-etal-2022-translation} was the first proposed to realize translation between textual description and molecular SMILES. BioT5 aims to enhance molecular understanding by incorporating protein. They also perform molecule captioning, which is equivalent to the inverse task of text-based molecule generation. (2) property-guided molecule generation is the inverse process of molecular property prediction, where molecules are generated based on specific biochemical property constraints. Notably, SPMM~\cite{chang2024bidirectional} was the first to establish a connection between 53 biochemical properties and SMILES sequences, making multi-constraint molecule generation possible. However, few existing models can simultaneously perform text-based or multi-constraint molecule generation and molecule captioning.

\noindent \textbf{Molecular Property Prediction.}
Deep learning models have been developed for molecular property prediction each with their own advantages and limitations. Transformer-based models design attention mechanism to capture contextual contexts from large-scale SMILES sequences~\cite{ross2022large}. The molecular graph can be directly obtained from SMILES sequences via RDKit~\cite{rdkit}. Graph-based models develop diverse graph neural networks to learn differentiable representations~\cite{wang2022molecular}. However, these methods ignore the potential that incorporating textual knowledge enables to realize new drug design objectives. Recently, a novel molecular pre-trained model named SPMM ~\cite{chang2024bidirectional} that extends the application of multimodal pre-training approaches by aligning molecular structures and biochemical properties. This paper extends the multimodal pre-training to patterns of text-sequence-property triplets, which is defined flexibly by LLM-understandable textual prompts.

\noindent \textbf{Instruction Tuning.}
Constructing specialized instruction datasets is an effective way to enable LLMs to better perform molecular-related tasks. For instance, Mol-Instructions~\cite{fang2023mol} provides a large-scale biomolecular dataset tailored for LLMs, covering a variety of instructions involving small molecules, proteins, and biomolecular texts.
ChemDFM~\cite{zhao2025developing} advances this paradigm by creating a broader dataset spanning molecular structures, reactions, and properties. Its two-stage training—domain pretraining followed by instruction tuning—enhances the model’s chemical understanding and reasoning capabilities.
More recently, GeLLM3O~\cite{dey2025mathtt} introduced MuMOInstruct, a high-quality dataset focused on multi-property molecule optimization, demonstrating strong generalization across diverse tasks.
Despite these advances, generating reliable and scalable molecular instruction data remains a key challenge, particularly for open-source models.
\section{Method}
\subsection{Overview of PEIT Framework}


The overview of PEIT framework is shown in Figure~\ref{figure:framework} (left), which consists of PEIT-GEN and PEIT-LLM. In PEIT-GEN, we generate a large number of ``SMILES-text'' and ``SMILES-property'' pairs to serve as multi-modal data. Then we design multiple multi-modal alignment objectives  to pre-train PEIT-GEN. In PEIT-LLM, by using the pre-trained PEIT-GEN, we can predict a large number of triplets to generate more diverse SMILES inputs, and then construct diverse instruction data based on template filling. By utilizing the synthesized instruction data, PEIT-LLM enables the supervised fine-tuning of open-source LLMs including LLaMa~\cite{dubey2024llama} and  Qwen~\cite{Qwen2TechnicalReport}, enhancing the capabilities for multi-task molecule generation.


\subsection{Pre-training of PEIT-GEN}
\label{sec:peit-gen}
The pre-training stage of PEIT-GEN is shown in the right of  Figure~\ref{figure:framework}. 
For a given molecule, different representations \textcolor{black}{offer} unique and complementary features, which are crucial for comprehensive molecule understanding. 
PEIT-GEN aims to integrate information from three modalities simultaneously, including textual information $\mathcal{T}$ (text), molecular structure $\mathcal{S}$ (SMILES), and biochemical properties $\mathcal{P}$ (property-value). Such ability can help synthesizing sufficient instruction data for further \textcolor{black}{enhancing} the ability of LLMs. In particular, PEIT-GEN consists of three Transformer encoders ${\rm Enc}^t$, ${\rm Enc}^s$, ${\rm Enc}^p$ and two decoders ${\rm Dec}^t$, ${\rm Dec}^p$, and we design different training objectives to align features from different modalities.

\noindent\textbf{Cross-modal Representation Matching.} Following SPMM~\cite{chang2024bidirectional}, we leverage pre-trained models SciBERT~\cite{Beltagy2019SciBERT} as trainable \textcolor{black}{$\text{Enc}^t$} for encoding textual data,  BERT~\cite{devlin2019bert} as \textcolor{black}{$\text{Enc}^s$ and $\text{Enc}^p$} for encoding SMILES and properties. Then we obtain feature representations across all three modalities, establishing the foundation for feature alignment.

We propose cross-modal representation matching to align the representations from different perspectives by the same molecule. In particular, we \textcolor{black}{introduce the} SMILES-text matching loss $\mathcal{L}^{st}_{\text{match}}$ \textcolor{black}{and the} SMILES-property matching loss $\mathcal{L}^{sp}_{\text{match}}$, which \textcolor{black}{serve} as objectives for training \textcolor{black}{the encoders}. In this way, the model can effectively learn cross-modal relationships and improve performance in \textcolor{black}{multi-modal} tasks by aligning the feature spaces. The matching loss is calculated as follows:
\begin{equation}
       \scalebox{1.0}{$
        \mathcal{L}^{st}_{\text{match}} = \ell_{\text{CE}}\left(y^{st}_{\text{match}}, {\rm MLP}({\rm Enc}^s(\mathcal{S})\oplus {\rm Enc}^t(\mathcal{T}))\right)$,}
\end{equation}
\begin{equation}
\scalebox{1.0}{$
    \mathcal{L}^{sp}_{\text{match}} = \ell_{\text{CE}}\left(y^{sp}_{\text{match}}, {\rm MLP}({\rm Enc}^s(\mathcal{S})\oplus {\rm Enc}^p(\mathcal{P}))\right),$}
\end{equation}
where $y^{st}_{\text{match}}$ and $y^{sp}_{\text{match}}$ are labels as 0 or 1, indicating whether the corresponding SMILES-text or SMILES-property pairs are matching. ${\rm Enc}(\cdot)$ indicates the representation of the data (i.e., [CLS] token of Transformer encoder), $\oplus$ is the concatenation operation, and ${\rm MLP}$($\cdot$) is the trainable multi-layer perception. The encoders are optimized by the cross-entropy loss $\ell_{\text{CE}}$ using the given data from different modalities.


\noindent\textbf{Multi-modal Contrastive Learning.} The representation matching can be viewed as an explicit 2-way classification training. We further utilize contrastive learning to directly enhancing the representation by pulling semantically close neighbors together and pushing apart non-neighbors from data of different modalities. To calculate the similarity between the encoded features of different modalities, we extract the encoded features and then compute the instance-level similarities through the inner product:
\begin{equation}
\scalebox{1.0}{$\text{sim}(\mathcal{S}, \mathcal{T}) = \left ({\rm MLP}^s({\rm Enc}^s(\mathcal{S}))\right )^\mathsf{T} {\rm MLP}^t({\rm Enc}^t(\mathcal{T})),$}
\end{equation}
\begin{equation}
\scalebox{1.0}{$
\text{sim}(\mathcal{S}, \mathcal{P}) = \left ({\rm MLP}^s({\rm Enc}^s(\mathcal{S}))\right )^\mathsf{T} {\rm MLP}^p({\rm Enc}^p(\mathcal{P})),$}
\end{equation}
where ${\rm MLP}^s$, ${\rm MLP}^t$ and ${\rm MLP}^p$ are multi-layer perceptions applied to SMILES, text, and property representations, respectively. Then, for the given SMILES $\mathcal{S}$, text $\mathcal{T}$, and property $\mathcal{P}$, we compute the cross-modal batch-level similarities as follows:
\begin{equation}
\scalebox{1.2}{$
s_{s2t} = \frac{\exp(\text{sim}(\mathcal{S}, \mathcal{T})/\tau)}{\sum_{i=1}^{M} \exp(\text{sim}(\mathcal{S},    \mathcal{T}_i)/\tau)},$}
\end{equation}
\begin{equation}
\scalebox{1.2}{$
s_{s2p} = \frac{\exp(\text{sim}(\mathcal{S}, \mathcal{P})/\tau)}{\sum_{i=1}^{N} \exp(\text{sim}(\mathcal{S}, \mathcal{P}_i)/\tau)},$}
\end{equation}
where $M$ and $N$ represent the total number of texts and property in the batch of data pairs\textcolor{black}{, respectively.} $\tau$ is the temperature controlling the sharpness of the similarity. The intra-modal similarities $s_{s2s}$, $s_{p2p}$, and $s_{t2t}$ can be computed in similar manners.

Based on the cross-modal and intra-modal batch-level similarities, the contrastive loss is formulated by calculating the cross-entropy according to one-hot encoded similarity vectors $y$, where the value is 1 for pairs derived from the same molecule or 0 for all other combinations:
\begin{equation}
\scalebox{1.05}{$
\begin{aligned}
\mathcal{L}^{st}_{\text{contrastive}} &= \frac{1}{2}( \ell_{\text{CE}}(y_{s2t}, s_{s2t}) + \ell_{\text{CE}}(y_{t2s}, s_{t2s}) \\
&\quad + \ell_{\text{CE}}(y_{s2s}, s_{s2s}) + \ell_{\text{CE}}(y_{t2t}, s_{t2t}) ),
\end{aligned}$}
\end{equation}
\begin{equation}
\scalebox{1}{$
\begin{aligned}
\mathcal{L}^{sp}_{\text{contrastive}} &= \frac{1}{2}( \ell_{\text{CE}}(y_{s2p}, s_{s2p}) + \ell_{\text{CE}}(y_{p2s}, s_{p2s})  \\
&\quad + \ell_{\text{CE}}(y_{s2s}, s_{s2s}) + \ell_{\text{CE}}(y_{p2p}, s_{p2p}) ).
\end{aligned}$}
\end{equation}

\begin{figure}[t!]
    \centering
    \includegraphics[width=0.77\linewidth]{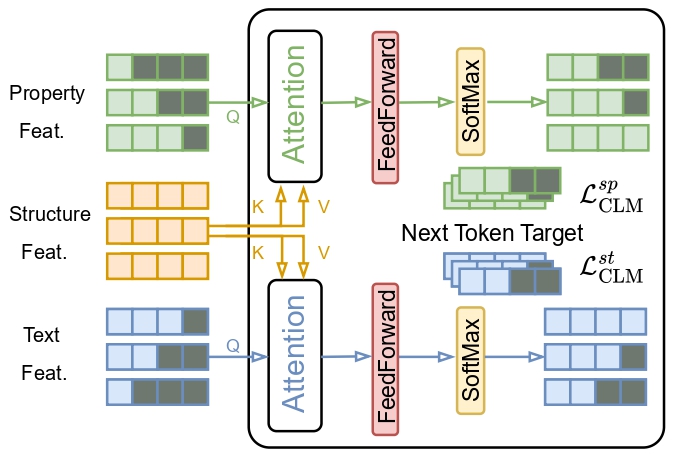}
    \caption{The cross-modal causal language modeling.}
    \label{causal}
\end{figure}

\noindent\textbf{Cross-modal Causal Language Modeling}. 
To further strengthen the model's capability in molecule captioning, we employ the causal language modeling (CLM) to enhance the model performance on text generation. Specifically, we design decoders to generate subsequent property and textual description sequences, under the guidance of SMILES features through cross-attention as show in Figure~\ref{causal}.




By introducing the SMILES features in attention layers for CLM training, the cross-modal CLM loss $\mathcal{L}^{st}_{\text{CLM}}$ and $\mathcal{L}^{sp}_{\text{CLM}}$ are computed as follows:
\begin{equation}
\scalebox{0.87}{$
\mathcal{L}^{st}_{\text{CLM}} = -\sum_{i=1}^{N} \sum_{j=1}^{n} \log {\rm Prob} \left(w_{j}^{(i)} \mid {\rm Dec}^t(\tilde{\mathbf{w}}_{: j}^{(i)}) ; \theta_{t}\right),
$}
\end{equation}
\begin{equation}
\scalebox{0.87}{$
\mathcal{L}^{sp}_{\text{CLM}} = -\sum_{i=1}^{N} \sum_{j=1}^{n} \log {\rm Prob} \left(w_{j}^{(i)} \mid {\rm Dec}^p(\tilde{\mathbf{w}}_{: j}^{(i)}) ; \theta_{p}\right),$}
\end{equation}
where ${\rm Prob}$ is the conditional probability to predict the word $w_{j}^{(i)}$ in the vocabulary, $N$ is the total number of samples, $n$ is the index of current words in each sample, $\tilde{\mathbf{w}}_{: j}^{(i)}$ is the sequence from begin to the $j$-th word in the $i$-th sample, $\theta_{t}$ and $\theta_{p}$ are the trainable parameters in two decoders. 


\noindent \textbf{Training.} The overall training objective for pre-training PEIT-GEN is to minimize the sum of all three types of losses across three modalities:
\begin{equation}
\scalebox{1}{$
\begin{aligned}
    \mathcal{L} &= \mathcal{L}^{st}_{\text{match}} + \mathcal{L}^{sp}_{\text{match}} + \alpha \mathcal{L}^{st}_{\text{contrastive}} + \alpha \mathcal{L}^{sp}_{\text{contrastive}} \\
    &\quad + \beta \mathcal{L}^{st}_{\text{CLM}} + \beta \mathcal{L}^{sp}_{\text{CLM}},
\end{aligned}
$}
\end{equation}
where we conducted a hyperparameter search for $\alpha$ and $\beta$, as shown in Table 1 in Appendix A. Based on the experimental results, we follow the setting in SPMM\cite{chang2024bidirectional} and adopt a ratio of $\alpha:\beta = 1:5$ to balance the loss terms.

\subsection{Instruction Tuning for PEIT-LLM}

\noindent \textbf{Template Filling.} The pre-trained PEIT-GEN offers unstructured data in the format of ``text-SMILES-properties'' (i.e., text-structure-property) triplets, which are stored in CSV files containing text, molecular structures, and information on 53 molecular biochemical properties. To obtain more task-specific data and to adapt to the strong instruction-following abilities of LLMs, we design templates for different downstream tasks, as shown in Figure 1 of Appendix B. For text-based molecule generation as example, we fix a general question format and then extract molecular descriptions from unstructured data to fill the pre-defined template, resulting in a natural question as instructions. The SMILES from unstructured triplets is used as the desired response. In this way, we can generate diverse task-specific instruction data in bulk for subsequent instruction-tuning.

\begin{table}[t!]
\centering
\scalebox{0.66}{
\begin{tabular}{lcccc}
\toprule
\textbf{Model} & \textbf{MC}\phantom{ (limited)} & \textbf{TBMG}\phantom{ (poor)} & \textbf{MPP}\phantom{ (poor)} & \textbf{MCMG}\phantom{ (poor)} \\
\midrule
MolT5 & \textcolor{teal}{\cmark}\phantom{ (limited)} & \textcolor{teal}{\cmark}\phantom{ (poor)} & \textcolor{red}{\xmark}\phantom{ (poor)} & \textcolor{red}{\xmark}\phantom{ (poor)} \\
BioT5 & \textcolor{teal}{\cmark}\phantom{ (limited)} & \textcolor{teal}{\cmark}\phantom{ (poor)} & \textcolor{red}{\xmark}\phantom{ (poor)} & \textcolor{red}{\xmark}\phantom{ (poor)} \\
MolXPT & \textcolor{teal}{\cmark}\phantom{ (limited)} & \textcolor{teal}{\cmark}\phantom{ (poor)} & \textcolor{red}{\xmark}\phantom{ (poor)} & \textcolor{red}{\xmark}\phantom{ (poor)} \\
Git-Mol & \textcolor{teal}{\cmark}\phantom{ (limited)} & \textcolor{teal}{\cmark}\phantom{ (poor)} & \textcolor{red}{\xmark}\phantom{ (poor)} & \textcolor{red}{\xmark}\phantom{ (poor)} \\
SPMM & \textcolor{red}{\xmark}\phantom{ (limited)} & \textcolor{red}{\xmark}\phantom{ (poor)} & \textcolor{teal}{\cmark}\phantom{ (poor)} & \textcolor{red}{\xmark}\phantom{ (poor)} \\
MolCA & \textcolor{teal}{\cmark}\phantom{ (limited)} & \textcolor{teal}{\cmark}\phantom{ (poor)} & \textcolor{red}{\xmark}\phantom{ (poor)} & \textcolor{red}{\xmark}\phantom{ (poor)} \\
Text+Chem-T5 & \textcolor{teal}{\cmark}\phantom{ (limited)} & \textcolor{teal}{\cmark}\phantom{ (poor)} & \textcolor{red}{\xmark}\phantom{ (poor)} & \textcolor{red}{\xmark}\phantom{ (poor)} \\
BioMedGPT & \textcolor{teal}{\cmark}\phantom{ (limited)} & \textcolor{red}{\xmark}\phantom{ (poor)} & \textcolor{red}{\xmark}\phantom{ (poor)} & \textcolor{red}{\xmark}\phantom{ (poor)} \\
InstructMol-GS & \textcolor{teal}{\cmark}\phantom{ (limited)} & \textcolor{red}{\xmark}\phantom{ (poor)} & \textcolor{red}{\xmark}\phantom{ (poor)} & \textcolor{red}{\xmark}\phantom { (poor)} \\
MolReGPT & \textcolor{teal}{\cmark}\phantom{ (limited)} & \textcolor{teal}{\cmark}\phantom{ (poor)} & \textcolor{red}{\xmark}\phantom{ (poor)} & \textcolor{red}{\xmark}\phantom{ (poor)} \\
Mol-Instructions & \textcolor{teal}{\cmark}\phantom{ (limited)} & \textcolor{teal}{\cmark}\phantom{ (poor)} & \textcolor{teal}{\cmark} (poor) & \textcolor{teal}{\cmark} (poor) \\
LLaMa, Qwen & \textcolor{teal}{\cmark} (limited) & \textcolor{teal}{\cmark} (poor) & \textcolor{teal}{\cmark} (poor) & \textcolor{teal}{\cmark} (poor) \\
\rowcolor{gray!20}
\textbf{PEIT-LLM (Ours)} & \textcolor{teal}{\cmark}\phantom{ (limited)} & \textcolor{teal}{\cmark}\phantom{ (poor)} & \textcolor{teal}{\cmark}\phantom{(poor)} & \textcolor{teal}{\cmark}\phantom{ (poor)} \\
\bottomrule
\end{tabular}}
\caption{Comparing PEIT-LLM with biomolecular models and LLMs on molecular-related tasks. MC: Molecule Captioning. TBMG: Text-Based Molecule Generation. MPP: Molecular Property Prediction. MCMG: Multi-Constraint Molecule Generation.}
\label{table:comparison}
\end{table}


\begin{table*}[t!]
    \centering
    \scalebox{0.78}{
    \begin{tabular}{lrcccccc}
        \toprule
        \textbf{Model} & \textbf{Data Size} ↓ & \textbf{BLEU-2} ↑ & \textbf{BLEU-4} ↑ & \textbf{METEOR} ↑ & \textbf{ROUGE-1} ↑ & \textbf{ROUGE-2} ↑ & \textbf{ROUGE-L} ↑ \\
        \midrule
        MolT5-small~\cite{edwards-etal-2022-translation} & 100M & 0.513 & 0.398 & 0.492 & 0.567 & 0.412 & 0.501 \\
        MolT5-large~\cite{edwards-etal-2022-translation} & 100M & 0.594 & 0.508 & 0.613 & 0.654 & 0.508 & 0.592 \\
        BioT5~\cite{pei2023biot5} & 33M & \textbf{0.635} & \textbf{0.556} & \underline{0.656} & \underline{0.692} & \underline{0.559} & \underline{0.633} \\

        MolXPT~\cite{liu2023molxpt}\dag & 30M & 0.594 & 0.505 & 0.626 & 0.660 & 0.511 & 0.597 \\
        MolCA$_\text{w/ Galac}$~\cite{liu2023molca} & 2.3M & 0.616 & 0.524 & 0.639 & 0.674 & 0.533 & 0.615 \\
        Text+Chem-T5$_\text{augm}$~\cite{christofidellis2023unifying} & 11.5M & \underline{0.625} & 0.529 & 0.648 & 0.682 & 0.543 & 0.622 \\
        GIT-Mol~\cite{liu2024git}\dag & \underline{4.8M} & 0.352 & 0.263 & 0.533 & 0.575 & 0.485 & 0.560 \\
        \rowcolor{gray!20}
        PEIT-GEN (Ours) & \textbf{0.48M} & 0.598 & \underline{0.534} & \textbf{0.676} & \textbf{0.700} & \textbf{0.582} & \textbf{0.653} \\
        \bottomrule
    \end{tabular}}
\caption{Results on CHEBI-20 dataset for molecule captioning with different pre-trained models. \dag: Reported from papers accordingly. The best results in each column are \textbf{in bold}, and the second-best results are \underline{underlined}.}
\label{comparison experiment}
\end{table*}

\noindent \textbf{Multi-constraint Molecule Generation Task.} 
Molecule generation often requires to be conducted under multiple constraints rather than a single condition.  In this work, we propose a new task to assess molecule generation through a variety of descriptors, by comparing the alignment between the generated molecules and specific criteria to evaluate the generative performance of LLMs. By using the large-scale unstructured data generated by PEIT-GEN, we can effectively synthesize sufficient data for evaluation. Specifically, we follow SPMM~\cite{chang2024bidirectional} given the vast number of molecular attributes
and the complex combinations thereof, analyzing their impact
on results across different molecular counts poses a signifi-
cant computational challenge due to resource limitations. To address this, we selected representative ADMET~\cite{fu2024admetlab} properties—BalabanJ, MolLogP, ExactMolWt, QED, and TPSA—that capture molecular topology, electronic characteristics, and steric effects, while exhibiting low mutual correlation. Based on template filling, the predicted multi-property values are used to construct data for multi-constraint molecule generation. We employ instruction tuning to guide the LLM in generating molecules, and use RDKit~\cite{rdkit} to calculate the actual property values. RMSE and R$^2$ are then used to compare these values against the constraints, enabling a systematic evaluation of the LLM's performance in multi-constraint molecule generation tasks.

\noindent \textbf{Supervised Fine-tuning.} 
We select LLaMa3.1-8B~\cite{dubey2024llama} and Qwen2.5-7B~\cite{Qwen2TechnicalReport} as base LLMs. We then perform standard SFT~\cite{ouyang2022training} by using the ``instruction-response'' pairs. In practice, we construct totally 1 million instruction data of four different tasks (i.e., molecule captioning, text-based molecule generation, property prediction, and multi-constraint molecule generation) from 200k unstructured ``text-SMILES-properties'' triplets obtained by PEIT-GEN.


\subsection{Comparing PEIT-LLM with Biomolecular Models and Large Language Models}
Table~\ref{table:comparison} shows a comparison of our PEIT-LLM with existing pre-trained models and general LLMs on multiple molecular generation tasks. For most of the pre-trained models such as MolT5 and BioT5, they focus on molecule captioning and text-based molecule generation, which can not \textcolor{black}{handle} property-related tasks. SPMM is a specialized model for property prediction. However, it lacks generation ability due to the lack of textual descriptions. Current LLMs such as LLaMa and Qwen show strong performance on general NLP-based tasks through conversations or instruction-following. However, these general LLMs still have limitations in tasks related to molecule generation due to a lack of molecular knowledge. In contrast, through fine-tuning on diverse instruction data with rich molecular knowledge, PEIT-LLM can perform multiple molecule generation tasks simultaneously.

\section{Experiments}

\subsection{Experimental Setup}

\noindent\textbf{Dataset.} 
During the pre-training of PEIT-GEN, we extract 480k molecular SMILES from the ZINC dataset~\cite{irwin2012zinc} and generate corresponding SMILES–text pairs using MolT5. Meanwhile, we compute 53 biochemical properties for each molecule using RDKit, forming 480k “text–SMILES–property” triplets for training. We then further fine-tune the model on the CHEBI-20~\cite{edwards2021text2mol} dataset to enhance its performance on human-annotated data.

For pre-training PEIT-LLM, we utilize the 200k tri-modal data generated by PEIT-GEN and employ template filling to generate 200k instruction data for each downstream task. For molecular property prediction, we select two biochemical properties with distinct differences for evaluation, generating 200k instruction data for each property. Finally, we obtain a total of 1000k instruction data across four tasks for SFT. Similar to PEIT-GEN, molecular property prediction tasks on PEIT-LLM can be validated by RDKit on CHEBI-20 dataset.

To evaluate PEIT-GEN and PEIT-LLM, we follow MolT5 by using CHEBI-20 and MoleculeNet dataset~\cite{wu2018moleculenet}, with the standard splition into training, validation, and test sets with an 8:1:1 ratio. All property values are verified via RDKit. See details in Appendix C.

\begin{table}[t!]
\centering
\scalebox{0.66}{
\begin{tabular}{lcccc}
\toprule
\textbf{Model} & \textbf{BBBP} & \textbf{BACE} & \textbf{Clintox} & \textbf{SIDER} \\
\midrule
D-MPNN ~\cite{yang2019analyzing}          & 71.0±0.3     & 80.9±0.6     & 90.6±0.6        & 57.0±0.7      \\
N-GramRF ~\cite{liu2019n}        & 69.7±0.6     & 77.9±1.5     & 77.5±4.0        & \underline{66.8±0.7}     \\
N-GramXGB ~\cite{liu2019n}       & 69.1±0.8     & 79.1±1.3     & 87.5±2.7        & 65.5±0.7      \\
PretrainGNN  ~\cite{hu2019strategies}    & 68.7±1.3     & 84.5±0.7     & 72.6±1.5        & 62.7±0.8      \\
GROVER\textsubscript{large} ~\cite{rong2020self}     & 69.5±0.1     & 81.0±1.4     & 76.2±3.7        & 65.4±0.1      \\
ChemRL-GEM ~\cite{fang2022geometry}      & 72.4±0.4     & \underline{85.6±1.1}     & 90.1±1.3        & \textbf{67.2±0.4}      \\
ChemBERTa ~\cite{ahmad2022chemberta}$^\dag$       & 72.8        & 79.9        & 56.3           & -             \\
MolFormer ~\cite{ross2022large}       & 73.6±0.8     & \textbf{86.3±0.6}     & \underline{91.2±1.4}        & 65.5±0.2      \\
SPMM ~\cite{chang2024bidirectional}           & \textbf{74.1±0.6}     & 82.9±0.3     & 90.7±0.5        & 63.6±0.5      \\
\rowcolor{gray!20}
PEIT-GEN (Ours)        & \underline{73.6±0.7}     & 81.6±0.5     & \textbf{91.2±0.7}        & 62.7±0.9      \\
\bottomrule
\end{tabular}
}
\caption{Results on MoleculeNet dataset for 2-way property prediction. \dag: The standard deviation and the results on SIDER are not reported in literature.}
\label{moleculenet}
\end{table}

\noindent\textbf{Baseline Models.}
We compare our model, PEIT-GEN and PEIT-LLM, against three types of baselines as follows:
\textit{Baselines on molecule caption} such as {MolT5}~\cite{edwards-etal-2022-translation}, {BioT5}~\cite{pei2023biot5},  {MolCA}~\cite{liu2023molca}, {Text+Chem-T5}~\cite{christofidellis2023unifying}, {GIT-Mol}~\cite{liu2024git}.
\textit{Baselines on moleucular property prediction} such as {SPMM}~\cite{chang2024bidirectional}, {D-MPNN}~\cite{yang2019analyzing}, {PretrainGNN}~\cite{hu2019strategies}, {GROVER}~\cite{rong2020self}, {ChemRL-GEM}~\cite{fang2022geometry}. 
\textit{Baselines of LLMs} such as  {LLaMa3}~\cite{touvron2023llama}, {GPT3.5-turbo}~\cite{achiam2023gpt}, {Mol-Instructions}~\cite{fang2023mol}, {Qwen2.5}~\cite{Qwen2TechnicalReport}, {BioMedGPT}~\cite{zhang2024generalist}, {ChatGLM}~\cite{glm2024chatglm}, \textcolor{black}{LLaSMol}~\cite{yu2024llasmol},{ChemDFM}~\cite{zhao2025developing}, InstructMol-GS~\cite{cao2025instructmol}, {Gemini}~\cite{comanici2025gemini},
Details and evaluation metric are in Appendix D and E, respectively.

\begin{table*}[t!]
\centering
\scalebox{0.82}{
\begin{tabular}{lccccccc}
\toprule
\textbf{Model} & \textbf{\#Params} & \textbf{BLEU-2} ↑ & \textbf{BLEU-4} ↑ & \textbf{METEOR} ↑ & \textbf{ROUGE-1} ↑ & \textbf{ROUGE-2} ↑ & \textbf{ROUGE-L} ↑ \\
\midrule
LLaMa3~\cite{touvron2023llama} & \phantom{0.}7B & 0.032 & 0.002 & 0.117 & 0.121 & 0.010 & 0.065 \\
LLaMa3.1~\cite{dubey2024llama} & \phantom{0.}8B & 0.042 & 0.004 & 0.121 & 0.140 & 0.019 & 0.095 \\
Qwen2.5~\cite{Qwen2TechnicalReport} & \phantom{0.}7B & 0.049 & 0.007 & 0.188 & 0.177 & 0.029 & 0.112 \\
GPT-3.5-turbo~\cite{achiam2023gpt} & \phantom{0.}N/A$^\dag$ &  0.103 &  0.050 &  0.161 &  0.261 &  0.088 & 0.204 \\
Mol-Instructions~\cite{fang2023mol} & \phantom{0.}8B & 0.217 & 0.143 & 0.254 & 0.337 & 0.196 & 0.291 \\
BioMedGPT~\cite{zhang2024generalist} &  \phantom{.}10B & 0.234 & 0.141 & 0.308 & 0.386 & 0.206 & 0.332 \\
InstructMol-GS~\cite{cao2025instructmol} &  \phantom{0.}7B & \underline{0.475} & \underline{0.371} & \underline{0.509} & 0.566 & 0.394 & 0.502 \\
MolReGPT~\cite{li2024empowering}& \phantom{0.}N/A$^\dag$ & \textbf{0.565} & \textbf{0.482} & \textbf{0.585} & \textbf{0.623} & \textbf{0.450} & \textbf{0.543} \\
\textcolor{black}{LLaSMol}~\cite{yu2024llasmol} & \phantom{0.}7B &  - &  - & 0.452 & -  & - &  - \\
ChemDFM~\cite{zhao2025developing} & \phantom{0.}13B &  0.321 &  0.265 & 0.402 & 0.490  & 0.374 &  0.483 \\
Gemini3~\cite{comanici2025gemini} & \phantom{0.}N/A$^\dag$ &  0.141 &  0.073 & 0.208 & 0.184  & 0.097&  0.276 \\
\rowcolor{gray!20}
PEIT-LLM-Qwen2.5 (Ours) & \phantom{0.}7B & 0.422 & 0.314 & 0.468 & 0.535 & 0.361 & 0.477 \\
\rowcolor{gray!20}
PEIT-LLM-LLaMa3.1 (Ours) & \phantom{0.}8B & 0.461 & 0.356 & 0.502& \underline{0.569} & \underline{0.396} & \underline{0.505} \\
\midrule
\textbf{Model} & \textbf{\#Params} & \textbf{BLEU} $\uparrow$ & \textbf{Validity} $\uparrow$ & \textbf{Levenshtein} $\downarrow$ & \textbf{MACCS FTS} $\uparrow$ & \textbf{Morgan FTS} $\uparrow$ & \textbf{RDKit FTS} $\uparrow$ \\
\midrule
LLaMa3~\cite{touvron2023llama} & \phantom{0.}7B & 0.261 & 0.330 & 45.788 & 0.372 & 0.127 & 0.213 \\
LLaMa3.1~\cite{dubey2024llama} & \phantom{0.}8B & 0.270 & 0.368 & 43.183 & 0.411 & 0.138 & 0.248 \\
Qwen2.5~\cite{Qwen2TechnicalReport} & \phantom{0.}7B & 0.217 & 0.245 & 50.550 & 0.403 & 0.110 & 0.276 \\
GPT-3.5-turbo~\cite{achiam2023gpt} & \phantom{0.}N/A$^\dag$ &   0.489 &  0.802 &  52.130 &  0.705 &  0.367 & 0.462 \\
Mol-Instructions~\cite{fang2023mol} & \phantom{0.}8B & 0.345 & \textbf{1.000} & 41.367 &  0.412 & 0.147 & 0.231 \\
MolReGPT~\cite{li2024empowering} & \phantom{0.}N/A$^\dag$ & 0.790 & 0.887 & 24.910 & \underline{0.847}  & \underline{0.624} & 0.708 \\
\textcolor{black}{LLaSMol}~\cite{yu2024llasmol} & \phantom{0.}7B &  - &  - & -& -  & 0.617 &  - \\
Gemini3~\cite{comanici2025gemini} & \phantom{0.}N/A$^\dag$ &  0.632 &  \textbf{1.000} & 29.782 & 0.793  & 0.569&  0.638 \\
\rowcolor{gray!20}
PEIT-LLM-Qwen2.5 (Ours) & \phantom{0.}7B & \underline{0.810} & 0.950 & \underline{21.133} & 0.832 & 0.619 & 0.735 \\
\rowcolor{gray!20}
PEIT-LLM-LLaMa3.1 (Ours) & \phantom{0.}8B & \textbf{0.836} & \underline{0.970} & \textbf{18.030} & \textbf{0.875} & \textbf{0.661} & \textbf{0.776} \\
\bottomrule
\end{tabular}}
\caption{Results on CHEBI-20 dataset for molecule captioning (top) and text-based molecule generation (bottom) tasks. $\dag$: MolReGPT is based on closed-source ChatGPT-3.5 and its parameter size remains unknown.}
\label{mol-caption-and-text}
\end{table*}

\noindent\textbf{Implementation Details.}
We pre-train PEIT-GEN for 20 epochs using a batch size of 16, temperature $\tau=0.07$, and momentum 0.995 with the AdamW optimizer~\cite{loshchilov2017decoupled}. Fine-tuning is then performed on the CHEBI-20 training set for 200 epochs with a learning rate of 5e-4. For supervised fine-tuning of PEIT-LLM, we use the LLaMa-Factory~\cite{zheng2024llamafactory} framework with LoRA~\cite{hu2022lora} for 6 epochs, a batch size of 3, and learning rate of 5e-5. The total parameter count of the three encoders and two decoders is 533.8M, with an additional 2.2M parameters for other components. Experiments are conducted on NVIDIA 4090 GPUs with 24GB memory.

\subsection{Comparing PEIT-GEN with Pre-trained Biomolecular Models}

\noindent\textbf{Molecule Captioning.}
Results on molecule captioning using CHEBI-20 dataset are shown in Table~\ref{comparison experiment}. Our model demonstrates superior performance in generating high-quality and relevant molecular caption. PEIT-GEN achieved the best results in METEOR and ROUGE, and the second-best performance in BLEU-4. Compared to BioT5 which performs the best in BLEU, our approach requires significantly less data. This indicates that using domain-specific models to generate paired data for pre-training is more efficient than single-modality pre-training.

\noindent\textbf{Molecular Property Prediction.} 
We evaluate the generalization ability of PEIT-GEN on the MoleculeNet benchmark~\cite{wu2018moleculenet} using four widely adopted tasks. As shown in Table~\ref{moleculenet}, PEIT-GEN outperforms specialized models such as MolFormer~\cite{ross2022large} and ChemRL-GEM~\cite{fang2022geometry} on the Clintox dataset. Despite using less pre-training data, it remains competitive on other subsets. To further demonstrate its predictive strength across 53 molecular properties, we present a relative difference analysis in Figure 2 of Appendix F, highlighting PEIT-GEN’s strong generalization in property prediction.

\subsection{Comparing PEIT-LLM with LLMs}

\noindent\textbf{Molecule Captioning.}
As shown in the top of Table~\ref{mol-caption-and-text}, the comparison results show that our model outperforms general-purpose Qwen-2.5 and LLaMa3.1 as well as Mol-Instructions and BioMedGPT, which were trained using a biochemical information instruction dataset for SFT. PEIT-LLM achieved the second-best performance on the ROUGE metric and demonstrated competitive results compared to InstructMol-GS, which was trained solely on the CHEBI-20 dataset and has a similar parameter scale as our base model.

\noindent\textbf{Text-based Molecule Generation.}
Results on the CHEBI-20 test set are presented at the bottom of Table~\ref{mol-caption-and-text}. PEIT-LLM outperforms all baselines on numerical metrics, including BLEU, Levenshtein Distance, and fingerprint similarities based on MACCS, Morgan, and RDKit. Although Mol-Instructions achieves the highest Validity score, the results demonstrate that PEIT-LLM, after multi-task instruction fine-tuning, effectively captures key molecular structures and corresponding textual representations. Case study in Table 5 of Appendix G further supports these findings and indirectly validates the quality of the data generated by PEIT-GEN.

\begin{table}[t!]
\centering
\scalebox{0.64}{
\begin{tabular}{lrrrr}
	    \toprule
        \multirow{2.2}*{\textbf{Model}}&\textbf{MolWt PP}&\textbf{MolLogP PP}&\multicolumn{2}{c}{\textbf{Five-Property CG}}\\
         \cmidrule(lr){2-2}\cmidrule(lr){3-3}\cmidrule(lr){4-5}
        &(RMSE) $\downarrow$&(RMSE) $\downarrow$&(RMSE) $\downarrow$&(R$^2$) $\uparrow$\\
      \midrule
     LLaMa3~\cite{touvron2023llama}   & 491.542 & 561.523 & 79.125 & -0.639 \\
LLaMa3.1~\cite{dubey2024llama}   & 544.517 & 552.521 & 74.646 & -0.652 \\
Qwen2.5~\cite{Qwen2TechnicalReport}    & 100.161 & 132.141 & 75.991 & -0.967 \\
Mol-Instructions~\cite{fang2023mol}   & 72.172 & 1.313 & 71.991 & -0.352 \\
\rowcolor{gray!20}
PEIT-LLM-Qwen2.5 (ours)   & \underline{14.164} & \underline{0.164} & \underline{19.750} & \underline{0.550} \\
\rowcolor{gray!20}
PEIT-LLM-LLaMa3.1 (ours)   & \textbf{13.918} & \textbf{0.141} & \textbf{14.212} & \textbf{0.613} \\
     \bottomrule
\end{tabular}
}
\caption{Results on property prediction (PP), and five-property constraint molecule generation (CG) with different LLMs.}
\label{combined_results1}
\end{table}

\noindent\textbf{Molecular Property Prediction.}
For single-property prediction, due to the large number of available properties, we select two representative examples: ExactMolWt, which typically has large numerical values (100 to 1000), and MolLogP, with smaller values (–5 to 10), as shown in Table~\ref{combined_results1}. The results show that PEIT-LLM consistently outperforms other LLMs in predicting these biochemical properties, demonstrating strong sensitivity and adaptability to molecular property scales. This highlights the effectiveness of multi-task SFT in enhancing LLMs’ understanding of molecular characteristics and further validates the quality and reliability of our molecular property instruction dataset. A case study is provided in Table 6 of Appendix G for further illustration.

\noindent\textbf{Multi-constraint Molecule Generation.} Results for our proposed multi-constraint molecule generation task is shown in Table~\ref{combined_results1}. PEIT-LLM surpasses baselines by large margin in both RMSE and R$^2$ metrics. \textcolor{black}{Case study is provided in Table 7 of Appendix G to further illustrate this point.} Note that this task requires the model to meet the demands of multiple properties with precise values, placing high demands on the model's overall understanding capability. General-purpose LLMs, or those not specifically trained for this task, lack the required information storage and fitting abilities. The model gain strong molecular understanding capabilities through property enhanced instruction tuning.

\begin{figure}[t!]
    \centering
    \includegraphics[scale=0.30]{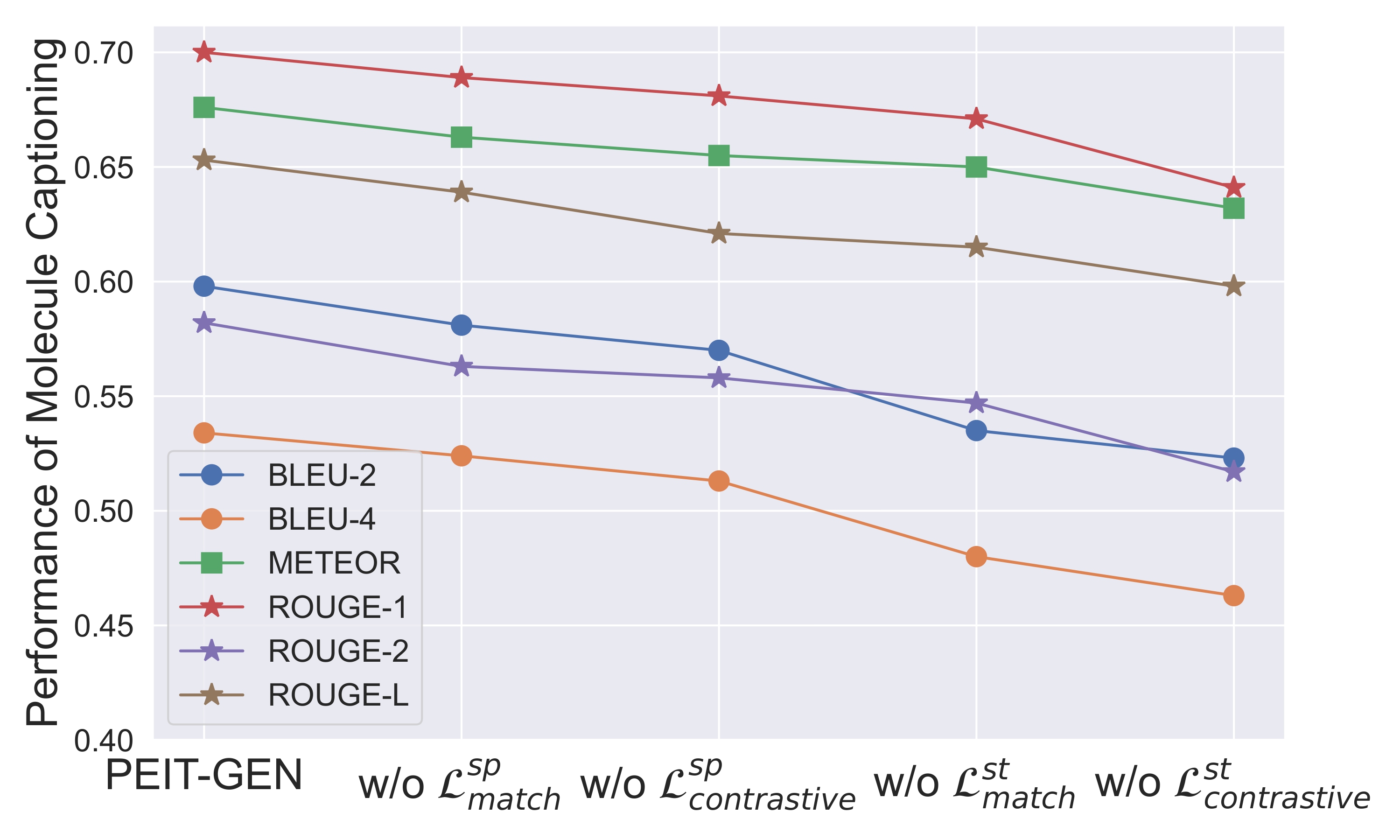}
    \caption{Ablation study on PEIT-GEN pre-training objectives $\mathcal{L}^{sp}_{\text{match}}$, $ \mathcal{L}^{st}_{\text{match}}$, $\mathcal{L}^{sp}_{\text{contrastive}}$, and $\mathcal{L}^{st}_{\text{contrastive}}$.}
    \label{ablation1}
\end{figure}

\begin{figure}[t!]
    \centering
    \includegraphics[width=0.88\linewidth]{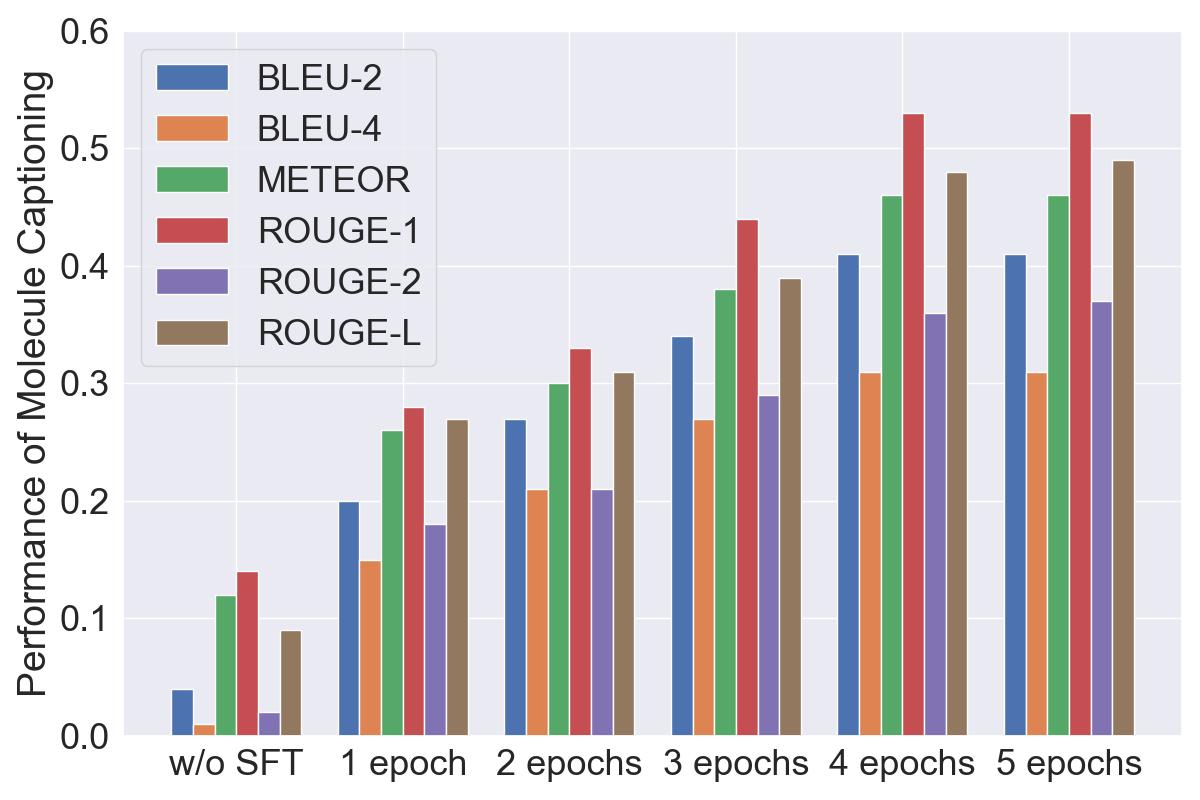}
    \caption{The impact of SFT steps on molecule captioning.}
    \label{ablation2}
\end{figure}

\begin{figure}[t!]
    \centering
    \includegraphics[width=0.88\linewidth]{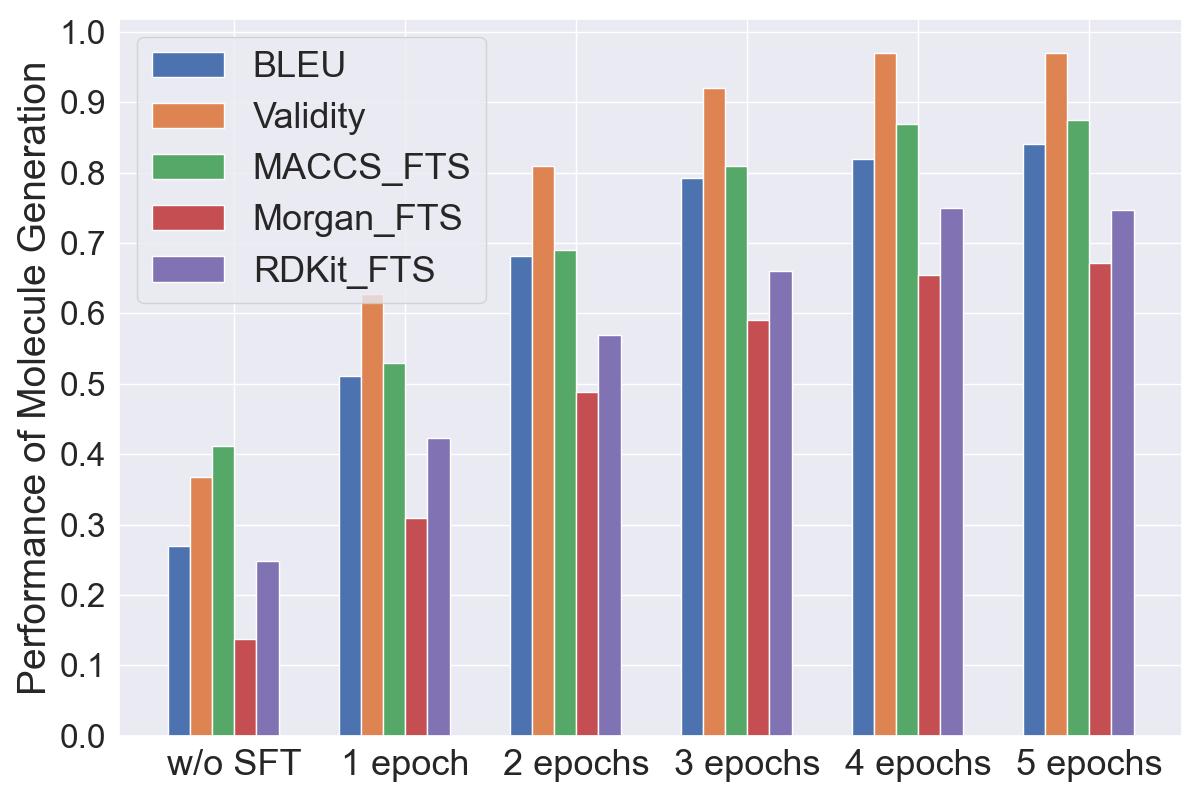}
    \caption{The impact of SFT steps on molecule generation.}
    \label{ablation4}
\end{figure}

\subsection{Analyses}
\noindent\textbf{Ablation Study.}\textcolor{black}{ ~Figure~\ref{ablation1}  presents an ablation study of the cross-modal matching loss $\mathcal{L}_{\text{match}}$ and cross-modal contrastive loss $\mathcal{L}_{\text{contrastive}}$in the PEIT-GEN model for the molecule captioning task ($\mathcal{L}^{st}_{\text{CLM}}$ and $\mathcal{L}^{sp}_{\text{CLM}}$ are necessary for generation via decoders, thus we do not consider them in ablation study). By removing these training objectives, the performance degradation across all metrics. This demonstrates that both $\mathcal{L}_{\text{match}}$ and $\mathcal{L}_{\text{contrastive}}$ are helpful in cross-modal feature alignment, thereby enhancing the performance of molecule captioning.} We further conducted an ablation study on the above loss components in the molecular property prediction task, as shown in the Table 3 of Appendix G.


\noindent\textbf{Impact of SFT steps.} Figure~\ref{ablation2} and Figure~\ref{ablation4} illustrate the outcomes of PEIT-LLM across various tasks with different SFT steps. We observe that the performance consistently improves during the initial epochs for all tasks, indicating that the instructional data is beneficial for each, where the performance tends to plateau around epochs 5-6. 


\noindent\textbf{Out-of-Distribution Evaluation.} To further evaluate the molecular understanding of PEIT-LLM on unseen data, we tested it on the Mol-Instructions~\cite{fang2023mol} test set, without using the full instructions for pre-training and fine-tuning PEIT-LLMs. As shown in Table~\ref{table:metrics}, PEIT-LLM outperforms all general-purpose LLMs as well as smaller domain-specific models highlighting the strong generalization ability of PEIT-LLM across diverse molecular instruction tasks.
\section{Conclusion}

\begin{table}[t!]
\centering
\resizebox{\linewidth}{!}{
\begin{tabular}{lcccc}
\hline
Model & \textbf{BLEU↑} & \textbf{METEOR↑} & \textbf{ROUGE-2↑} & \textbf{ROUGE-L↑}\\
\hline
Galactica-6.7B~\cite{taylor2022galactica} & 0.008 & 0.065 &  0.015 & 0.063 \\
MolT5-248M~\cite{edwards-etal-2022-translation} & 0.001 & 0.033 & 0.001 & 0.034 \\
Vicuna-7B~\cite{vicuna2023} & 0.011 & 0.168 & 0.055 & 0.130 \\
Text+Chem T5-223M~\cite{christofidellis2023unifying} & 0.036 & 0.139 & 0.075 & 0.119 \\
ChatGLM-6B~\cite{glm2024chatglm}& 0.011 & 0.105 &  0.066 & 0.148 \\
LLaMa3.1-8B~\cite{dubey2024llama} & 0.014 & 0.184 & 0.066 & 0.148 \\
Qwen2.5-7B~\cite{Qwen2TechnicalReport} & 0.009 & 0.169 & 0.047 & 0.119 \\
\rowcolor{gray!20}
PEIT-LLM-Qwen2.5-7B (Ours) & 0.051 & 0.208 & 0.121 & 0.178 \\
\rowcolor{gray!20}
PEIT-LLM-LLaMa3.1-8B (Ours) & \underline{0.053} & \underline{0.215} & \underline{0.125} & \underline{0.184} \\
\hdashline
\textcolor{gray}{Mol-Instructions-8B~\cite{fang2023mol}} & \textcolor{gray}{0.143} & \textcolor{gray}{0.254} & \textcolor{gray}{0.196} & \textcolor{gray}{0.291} \\
\hline
\end{tabular}
}
\caption{Out-of-distribution results on the molecule captioning task using Mol-Instructions~\protect\cite{fang2023mol} evaluation set.
\protect\textcolor{gray}{Mol-Instructions} denotes a fully supervised baseline trained with LLaMA3.1-8B using the entire training instructions, serving as an upper bound for SFT models.}
\label{table:metrics}
\end{table}

We propose PEIT, a framework that enables LLMs to perceive multi-modal features for multi-task molecule generation. PEIT aligns molecular structures, textual descriptions, and biochemical properties through multi-modal representation learning. It leverages templates to synthesize diverse, task-specific instruction data for LLMs. We also introduce a challenging multi-constraint molecule generation task, which requires generating novel molecules that satisfy multiple property constraints. Results show that PEIT outperforms various biomolecular models and LLMs on captioning, generation, and property prediction tasks.


\section*{Acknowledgments}
\textcolor{black}{The work is supported in part by the National Natural Science Foundation of China (62573372, 62425204, U22A2037, 62450002, 62432011), the Hunan Provincial Key Research and Development Program Project (2025JK2003), the Guangdong Basic and Applied Basic Research Foundation (2026A1515011358), the Shenzhen Natural Science Foundation (JCYJ20250604181610014), and the Fundamental and Interdisciplinary Disciplines Breakthrough Plan of the Ministry of Education of China (JYB2025XDXM602).}

\bibliographystyle{named}
\bibliography{ijcai26}

@article{achiam2023gpt,
  title={{GPT}-4 technical report},
  author={OpenAI},
  journal={arXiv preprint arXiv:2303.08774},
  year={2023},
  url={https://arxiv.org/abs/2303.08774}
}

@misc{vicuna2023,
  title={Vicuna: An Open-Source Chatbot Impressing GPT-4 with 90\%* ChatGPT Quality},
  author={Chiang, Zhihao and Zhu, Zhuohan and Zeng, Xinyang and others},
  year={2023},
  eprint={2304.05334},
  archivePrefix={arXiv},
  primaryClass={cs.CL}
}

@article{touvron2023llama,
  title={Llama: Open and efficient foundation language models},
  author={Touvron, Hugo and Lavril, Thibaut and Izacard, Gautier and Martinet, Xavier and Lachaux, Marie-Anne and Lacroix, Timoth{\'e}e and Rozi{\`e}re, Baptiste and Goyal, Naman and Hambro, Eric and Azhar, Faisal and others},
  journal={arXiv preprint arXiv:2302.13971},
  year={2023},
  url={https://arxiv.org/abs/2302.13971}
}

@article{dubey2024llama,
  title={The llama 3 herd of models},
  author={Dubey, Abhimanyu and Jauhri, Abhinav and others},
  journal={arXiv preprint arXiv:2407.21783},
  year={2024},
  url={https://arxiv.org/abs/2407.21783}
}

@article{petroni2019language,
  title={Language models as knowledge bases?},
  author={Petroni, Fabio and Rockt{\"a}schel, Tim and Lewis, Patrick and others},
  journal={arXiv preprint arXiv:1909.01066},
  year={2019},
  url={https://aclanthology.org/D19-1250/}
}

@article{zhang2023instruction,
  title={Instruction tuning for large language models: A survey},
  author={Zhang, Shengyu and Dong, Linfeng and Li, Xiaoya and Zhang, Sen and Sun, Xiaofei and Wang, Shuhe and Li, Jiwei and Hu, Runyi and Zhang, Tianwei and Wu, Fei and others},
  journal={arXiv preprint arXiv:2308.10792},
  year={2023},
  url={https://arxiv.org/abs/2308.10792}
}

@inproceedings{fang2023mol,
    author = {Fang, Yin and Liang, Xiaozhuan and others},
    title = {Mol-\uppercase{I}nstructions: A large-scale biomolecular instruction dataset for large language models},
    booktitle = {International Conference on Learning Representations},
    year = {2023},
    url={https://iclr.cc/media/iclr-2024/Slides/18554.pdf}
}

@article{zhavoronkov2018artificial,
  title={Artificial intelligence for drug discovery, biomarker development, and generation of novel chemistry},
  author={Zhavoronkov, Alex},
  journal={Molecular Pharmaceutics},
  volume={15},
  number={10},
  pages={4311--4313},
  year={2018},
  url={https://pubs.acs.org/doi/10.1021/acs.molpharmaceut.8b00930}
}

@article{elton2019deep,
  title={Deep learning for molecular design—a review of the state of the art},
  author={Elton, Daniel C and Boukouvalas, Zois and Fuge, Mark D and Chung, Peter W},
  journal={Molecular Systems Design \& Engineering},
  volume={4},
  number={4},
  pages={828--849},
  year={2019},
  url={https://arxiv.org/abs/1903.04388}
}

@article{grisoni2023chemical,
  title={Chemical language models for de novo drug design: Challenges and opportunities},
  author={Grisoni, Francesca},
  journal={Current Opinion in Structural Biology},
  volume={79},
  pages={102527},
  year={2023},
  url={https://www.sciencedirect.com/science/article/pii/S0959440X23000015}
}

@article{wu2018moleculenet,
  title={MoleculeNet: a benchmark for molecular machine learning},
  author={Wu, Zhenqin and Ramsundar, Bharath and Feinberg and others},
  journal={Chemical science},
  volume={9},
  number={2},
  pages={513--530},
  year={2018},
  url={https://pubs.rsc.org/en/content/articlelanding/2018/sc/c7sc02664a}
}

@article{yu2024llasmol,
    title={LlaSMol: Advancing Large Language Models for Chemistry with a Large-Scale, Comprehensive, High-Quality Instruction Tuning Dataset},
    author={Botao Yu and Frazier N. Baker and Ziqi Chen and Xia Ning and Huan Sun},
    journal={arXiv preprint arXiv:2402.09391},
    year={2024}
}

@inproceedings{edwards-etal-2022-translation,
    title = {Translation between Molecules and Natural Language},
    author = {Edwards, Carl and Lai, Tuan and
      others},
    booktitle = {Proceedings of the 2022 Conference on Empirical Methods in Natural Language Processing},
    year = {2022},
    pages = {375--413},
    url={https://aclanthology.org/2022.emnlp-main.26/}
}

@article{chang2024bidirectional,
  title={Bidirectional generation of structure and properties through a single molecular foundation model},
  author={Chang, Jinho and Ye, Jong Chul},
  journal={Nature Communications},
  volume={15},
  number={1},
  pages={2323},
  year={2024},
  url={https://www.nature.com/articles/s41467-024-46440-3}
}

@inproceedings{pei2023biot5,
    title = "{B}io{T}5: Enriching Cross-modal Integration in Biology with Chemical Knowledge and Natural Language Associations",
    author = "Pei, Qizhi  and
      Zhang, Wei  and
      others",
    booktitle = "Proceedings of the 2023 Conference on Empirical Methods in Natural Language Processing",
    year = "2023",
    pages = "1102--1123",
    url={https://aclanthology.org/2023.emnlp-main.70/}
}

@article{liu2024git,
  title={Git-mol: A multi-modal large language model for molecular science with graph, image, and text},
  author={Liu, Pengfei and Ren, Yiming and Tao, Jun and Ren, Zhixiang},
  journal={Computers in biology and medicine},
  volume={171},
  pages={108073},
  year={2024},
  url={https://www.sciencedirect.com/science/article/pii/S0010482524001574}
}

@inproceedings{liu2023molxpt,
    title = "{M}ol{XPT}: Wrapping Molecules with Text for Generative Pre-training",
    author = "Liu, Zequn  and
      Zhang, Wei  and
      Xia, Yingce  and
      others",
    booktitle = "Proceedings of the 61st Annual Meeting of the Association for Computational Linguistics (Volume 2: Short Papers)",
    year = "2023",
    pages = "1606--1616",
    url={https://aclanthology.org/2023.acl-short.138/}
}

@article{Qwen2TechnicalReport,
  title={Qwen2 technical report},
  author={Yang, An and Yang, Baosong and Hui, Binyuan and Zheng, Bo and Yu, Bowen and Zhou, Chang and Li, Chengpeng and Li, Chengyuan and Liu, Dayiheng and Huang, Fei and others},
  journal={arXiv preprint arXiv:2407.10671},
  year={2024},
  url={https://arxiv.org/abs/2407.10671}
}

@article{ross2022large,
  title={Large-scale chemical language representations capture molecular structure and properties},
  author={Ross, Jerret and Belgodere, Brian and Chenthamarakshan, Vijil and Padhi, Inkit and Mroueh, Youssef and Das, Payel},
  journal={Nature Machine Intelligence},
  volume={4},
  number={12},
  pages={1256--1264},
  year={2022},
  url={https://www.nature.com/articles/s42256-022-00580-7}
}

@article{rdkit,
  title={RDKit: A software suite for cheminformatics, computational chemistry, and predictive modeling},
  author={Landrum, Greg and others},
  journal={Greg Landrum},
  volume={8},
  number={31.10},
  pages={5281},
  year={2013},
  url={https://www.rdkit.org/RDKit_Overview.pdf}
}

@article{wang2022molecular,
  title={Molecular contrastive learning of representations via graph neural networks},
  author={Wang, Yuyang and Wang, Jianren and Cao, Zhonglin and Barati Farimani, Amir},
  journal={Nature Machine Intelligence},
  volume={4},
  number={3},
  pages={279--287},
  year={2022},
  url={https://www.nature.com/articles/s42256-022-00447-x}
}

@inproceedings{Beltagy2019SciBERT,
    title = "{S}ci{BERT}: A Pretrained Language Model for Scientific Text",
    author = "Beltagy, Iz  and
      Lo, Kyle  and
      Cohan, Arman",
    booktitle = "Proceedings of the 2019 Conference on Empirical Methods in Natural Language Processing",
    year = "2019",
    pages = "3615--3620",
    url={https://aclanthology.org/D19-1371.pdf}
}

@inproceedings{devlin2019bert,
    title = "{BERT}: Pre-training of Deep Bidirectional Transformers for Language Understanding",
    author = "Devlin, Jacob  and
      Chang, Ming-Wei and others",
    booktitle = "Proceedings of the 2019 Conference of the North {A}merican Chapter of the Association for Computational Linguistics: Human Language Technologies",
    year = "2019",
    pages = "4171--4186",
    url={https://aclanthology.org/N19-1423/}
}

@inproceedings{ouyang2022training,
author = {Ouyang, Long and Wu, Jeff and others},
title = {Training language models to follow instructions with human feedback},
year = {2024},
booktitle = {Proceedings of the 36th International Conference on Neural Information Processing Systems},
numpages = {15},
url={https://dl.acm.org/doi/10.5555/3600270.3602281}
}

@inproceedings{loshchilov2017decoupled,
    author = {Loshchilov, I},
    title = {Decoupled weight decay regularization},
    booktitle = {International Conference on Learning Representations},
    year = {2017},
    url={https://arxiv.org/abs/1711.05101}
}

@article{zheng2024llamafactory,
  title={Llamafactory: Unified efficient fine-tuning of 100+ language models},
  author={Zheng, Yaowei and Zhang, Richong and Zhang, Junhao and Ye, Yanhan and Luo, Zheyan and Feng, Zhangchi and Ma, Yongqiang},
  journal={arXiv preprint arXiv:2403.13372},
  year={2024},
  url={https://arxiv.org/abs/2403.13372}
}

@inproceedings{hu2022lora,
    title={Lo{RA}: Low-Rank Adaptation of Large Language Models},
    author={Edward J Hu and Yelong Shen and Phillip Wallis and Zeyuan Allen-Zhu and Yuanzhi Li and Shean Wang and Lu Wang and Weizhu Chen},
    booktitle={International Conference on Learning Representations},
    year={2022},
    url={https://iclr.cc/virtual/2022/poster/6319}
}

@article{taylor2022galactica,
  title={Galactica: A large language model for science},
  author={Taylor, Ross and Kardas, Marcin and Cucurull, Guillem and Scialom, Thomas and Hartshorn, Anthony and Saravia, Elvis and Poulton, Andrew and Kerkez, Viktor and Stojnic, Robert},
  journal={arXiv preprint arXiv:2211.09085},
  year={2022}
}

@article{glm2024chatglm,
  title={Chatglm: A family of large language models from glm-130b to glm-4 all tools},
  author={GLM, Team and Zeng, Aohan and Xu, Bin and Wang, Bowen and Zhang, Chenhui and Yin, Da and Zhang, Dan and Rojas, Diego and Feng, Guanyu and Zhao, Hanlin and others},
  journal={arXiv preprint arXiv:2406.12793},
  year={2024}
}

@article{dey2025mathtt,
  title={{\texttt{GeLLM}$^{3O}$}: Generalizing Large Language Models for Multi-property Molecule Optimization},
  author={Dey, Vishal and Hu, Xiao and Ning, Xia},
  journal={arXiv preprint arXiv:2502.13398},
  url={https://arxiv.org/abs/2502.13398},
  year={2025}
}

@article{zhao2025developing,
  title={Developing ChemDFM as a large language foundation model for chemistry},
  author={Zhao, Zihan and Ma, Da and Chen, Lu and Sun, Liangtai and Li, Zihao and Xia, Yi and Chen, Bo and Xu, Hongshen and Zhu, Zichen and Zhu, Su and others},
  journal={Cell Reports Physical Science},
  volume={6},
  number={4},
  year={2025},
  publisher={Elsevier}
}

@article{fu2024admetlab,
  title={ADMETlab 3.0: an updated comprehensive online ADMET prediction platform enhanced with broader coverage, improved performance, API functionality and decision support},
  author={Fu, Li and Shi, Shaohua and Yi, Jiacai and others},
  journal={Nucleic acids research},
  volume={52},
  number={W1},
  pages={W422--W431},
  year={2024},
  publisher={Oxford University Press}
}

@article{irwin2012zinc,
  title={ZINC: a free tool to discover chemistry for biology},
  author={Irwin, John J and Sterling, Teague and Mysinger, Michael M and Bolstad, Erin S and Coleman, Ryan G},
  journal={Journal of chemical information and modeling},
  volume={52},
  number={7},
  pages={1757--1768},
  year={2012},
  publisher={ACS Publications},
  url={https://pubs.acs.org/doi/10.1021/ci3001277}
}

@inproceedings{edwards2021text2mol,
  title={Text2mol: Cross-modal molecule retrieval with natural language queries},
  author={Edwards, Carl and Zhai, ChengXiang and Ji, Heng},
  booktitle={Proceedings of the 2021 Conference on Empirical Methods in Natural Language Processing},
  pages={595--607},
  year={2021}
}

@article{yang2019analyzing,
  title={Analyzing learned molecular representations for property prediction},
  author={Yang, K. and others},
  journal={Journal of Chemical Information and Modeling},
  volume={59},
  number={8},
  pages={3370--3388},
  year={2019},
  publisher={American Chemical Society},
  url={https://pubs.acs.org/doi/10.1021/acs.jcim.9b00237}
}

@article{liu2019n,
  title={N-gram graph: Simple unsupervised representation for graphs, with applications to molecules},
  author={Liu, Shengchao and Demirel and others},
  journal={Advances in neural information processing systems},
  volume={32},
  year={2019},
  url={https://proceedings.neurips.cc/paper_files/paper/2019/file/2f3926f0a9613f3c3cc21d52a3cdb4d9-Paper.pdf}
}

@article{hu2019strategies,
  title={Strategies for pre-training graph neural networks},
  author={Hu, Weihua and Liu, Bowen and Gomes, Joseph and Zitnik, Marinka and Liang, Percy and Pande, Vijay and Leskovec, Jure},
  journal={arXiv preprint arXiv:1905.12265},
  year={2019},
  url={https://arxiv.org/abs/1905.12265}
}

@article{rong2020self,
  title={Self-supervised graph transformer on large-scale molecular data},
  author={Rong, Yu and Bian, Yatao and Xu, Tingyang and Xie, Weiyang and Wei, Ying and Huang, Wenbing and Huang, Junzhou},
  journal={Advances in neural information processing systems},
  volume={33},
  pages={12559--12571},
  year={2020},
  url={https://dl.acm.org/doi/abs/10.5555/3495724.3496777}
}

@article{fang2022geometry,
  title={Geometry-enhanced molecular representation learning for property prediction},
  author={Fang, Xiaomin and Liu, Lihang and Lei, Jieqiong and He, Donglong and Zhang, Shanzhuo and Zhou, Jingbo and Wang, Fan and Wu, Hua and Wang, Haifeng},
  journal={Nature Machine Intelligence},
  volume={4},
  number={2},
  pages={127--134},
  year={2022},
  publisher={Nature Publishing Group},
  url={https://www.nature.com/articles/s42256-021-00438-4}
}

@article{ahmad2022chemberta,
  title={Chemberta-2: Towards chemical foundation models},
  author={Ahmad, Walid and Simon, Elana and Chithrananda, Seyone and Grand, Gabriel and Ramsundar, Bharath},
  journal={arXiv preprint arXiv:2209.01712},
  year={2022},
  url={https://arxiv.org/abs/2209.01712}
}

@article{zhang2024generalist,
  title={A generalist vision--language foundation model for diverse biomedical tasks},
  author={Zhang, Kai and Zhou, Rong and Adhikarla, Eashan and Yan, Zhiling and Liu, Yixin and Yu, Jun and Liu, Zhengliang and Chen, Xun and Davison, Brian D and Ren, Hui and others},
  journal={Nature Medicine},
  pages={1--13},
  year={2024},
  publisher={Nature Publishing Group US New York},
  url={https://www.nature.com/articles/s41591-024-03185-2}
}

@inproceedings{cao2025instructmol,
  title={Instructmol: Multi-modal integration for building a versatile and reliable molecular assistant in drug discovery},
  author={Cao, He and Liu, Zijing and Lu, Xingyu and Yao, Yuan and Li, Yu},
  booktitle={Proceedings of the 31st International Conference on Computational Linguistics},
  pages={354--379},
  year={2025}
}

@article{li2024empowering,
  title={Empowering molecule discovery for molecule-caption translation with large language models: A chatgpt perspective},
  author={Li, Jiatong and Liu, Yunqing and Fan, Wenqi and Wei, Xiao-Yong and Liu, Hui and Tang, Jiliang and Li, Qing},
  journal={IEEE transactions on knowledge and data engineering},
  year={2024},
  publisher={IEEE},
  url={https://ieeexplore.ieee.org/abstract/document/10516270}

}

@inproceedings{liu2023molca,
    title={MolCA: Molecular Graph-Language Modeling with Cross-Modal Projector and Uni-Modal Adapter},
    author={Liu, Zhiyuan and Li, Sihang and others},
    booktitle={EMNLP},
    year={2023},
    url={https://openreview.net/forum?id=14WRhMNq7H}
}

@inproceedings{christofidellis2023unifying,
  title = 	 {Unifying Molecular and Textual Representations via Multi-task Language Modelling},
  author =       {Christofidellis, Dimitrios and Giannone, Giorgio and Born, Jannis and Winther, Ole and Laino, Teodoro and Manica, Matteo},
  booktitle = 	 {Proceedings of the 40th International Conference on Machine Learning},
  pages = 	 {6140--6157},
  year = 	 {2023},
  url = 	 {https://proceedings.mlr.press/v202/christofidellis23a.html},
}

@article{comanici2025gemini,
  title={Gemini 2.5: Pushing the frontier with advanced reasoning, multimodality, long context, and next generation agentic capabilities},
  author={Comanici, Gheorghe and Bieber, Eric and Schaekermann, Mike and Pasupat, Ice and Sachdeva, Noveen and Dhillon, Inderjit and Blistein, Marcel and Ram, Ori and Zhang, Dan and Rosen, Evan and others},
  journal={arXiv preprint arXiv:2507.06261},
  year={2025}
}

@article{guo2025deepseek,
  title={Deepseek-r1: Incentivizing reasoning capability in llms via reinforcement learning},
  author={Guo, Daya and Yang, Dejian and Zhang, Haowei and Song, Junxiao and Zhang, Ruoyu and Xu, Runxin and Zhu, Qihao and Ma, Shirong and Wang, Peiyi and Bi, Xiao and others},
  journal={arXiv preprint arXiv:2501.12948},
  year={2025}
}

\end{document}